\documentclass[sigconf]{acmart}
\AtBeginDocument{%
  }

\setcopyright{acmlicensed}
\copyrightyear{2025}
\acmYear{2025}
\acmDOI{XXXXXXX.XXXXXXX}
\acmConference[Conference acronym 'XX]{Make sure to enter the correct
  conference title from your rights confirmation email}{June 03--05,
  2025}{Woodstock, NY}
\acmISBN{978-1-4503-XXXX-X/2018/06}

\usepackage{subcaption}

\usepackage{xspace}

\PassOptionsToPackage{table}{xcolor}
\usepackage{multirow}
\usepackage{graphicx}
\usepackage{tabularx,booktabs}
\usepackage{rotating}
\usepackage{hyperref}

\usepackage{cleveref}
\crefformat{section}{\S#2#1#3}
\crefformat{subsection}{\S#2#1#3}
\crefformat{subsubsection}{\S#2#1#3}
\crefrangeformat{section}{\S\S#3#1#4 to~#5#2#6}
\crefmultiformat{section}{\S\S#2#1#3}{ and~#2#1#3}{, #2#1#3}{ and~#2#1#3}
\Crefformat{figure}{#2Fig.~#1#3}
\Crefmultiformat{figure}{Figs.~#2#1#3}{ and~#2#1#3}{, #2#1#3}{ and~#2#1#3}
\Crefformat{table}{#2Tab.~#1#3}
\Crefmultiformat{table}{Tabs.~#2#1#3}{ and~#2#1#3}{, #2#1#3}{ and~#2#1#3}
\Crefformat{appendix}{Appx.~\S#2#1#3}
\crefformat{algorithm}{Alg.~#2#1#3}
\crefformat{equation}{Eq.~#2#1#3}

\newcommand{\tableprompt}{
    \begin{table}[t]
        \small
        \centering
        \begin{tabular}{p{\dimexpr \linewidth-2\tabcolsep}}
        
            \toprule
            \textit{Table}: planet\_osm\_point \\
    \textit{Columns}: \\
    - name: Text (Location name) \\
    - osm\_id: Integer (Unique ID for the location or amenity) \\
    - amenity: Text (Type of amenity, e.g., "health", "school") \\
    - way: Geometry (Geometric point/line data representing the location) \\
    \midrule
    \textit{Notes}: \\
    - The database stores geographic information.\\
    - The "way" column stores location details, which can be queried using PostGIS functions for geographic operations. \\
    \midrule
    Generate a SQL query to answer the question: "\{question\}"\\
    \midrule
    \textit{Context}: "\{context\}"\\
    OUTPUT ONLY THE SQL QUERY:  \\
        \bottomrule
        \end{tabular}
        \caption{Sample LLM prompt for Text-to-SQL Setup. The \textit{context} is structured as follows: ``The OSMID of <location\_name> is <osm\_id>''}
        \label{tab:ttsql_prompt}
    \end{table}
}

\newcommand{\tabledataset}{

\begin{table*}[t]
\centering
\begin{tabular}{c|p{4.4cm}|p{4.4cm}|p{2.8cm}|p{1.6cm} | p{2cm}}
\toprule
 & \textbf{Template} & \textbf{Question} & \textbf{Answer} & \textbf{Category} & \textbf{Concepts}  \\
\midrule
1 & What [amenity] is adjacent to [location]? & What restaurant is adjacent to Luther Burbank Savings? & Domenico’s & Geo-entity & Amenity type \& Closeness \\
\hline
2 & What are the [amenity] within 50m of [location]? & What are the bars within 50m of South Park Brewing? & Hamilton’s Tavern & Geo-entity & Amenity type \&  Distance  \\
\hline
3 & What amenities are within a 100m radius from [location]?  & What amenities are within a 100m radius of Yorba Rose? & Fuel & Attribute & Amenity type \&  Distance  \\
\hline
4 & What amenity is available at [location]? & What amenity is available at Port Police? & Police & Attribute & Amenity type \\
\hline
5 & Which is closer to [location A], [location B] or [location C]? & Which is closer to ChargePoint, Chicken Maison or Ospi? & Chicken Maison & Geo-entity & Comparison \&  Closeness  \\
\hline
6 & What is the nearest [amenity] [direction] of [location]? & What is the nearest snack\_cart west of Colours Wheelchairs? & Tomorrowland & Geo-entity & Amenity type \&  Direction\\
\hline
7 & What is the nearest [amenity] to [location]? & What is the nearest clinic to East Wing? & Garden's Elderly Care & Geo-entity & Amenity type \&  Compare \\
\midrule
\midrule
8 &  Which [amenity] is nearest to the junction of [location A] and [location B]? & Which driving\_school is nearest to the junction of 10th Street and 10th Street West? & Freeway Easy Traffic and Driving School & Geo-entity & Amenity type \& Intersect \\
\hline
9 & How far is [location A] from [location B]? & How far is 119th Street \& Laflin from 100 Forest Place? & 35141 meters & Distance & Distance\\ 

\bottomrule
\end{tabular}
\caption{Question templates and example question-answer pairs. The first seven question types are solvable by both retrieval-based and LLM-based approaches, while the last two types are solvable exclusively by LLM-based methods. The ``Category'' column shows the answer category, and the ``Concepts'' column explains the spatial concepts involved in the question. 
}
\label{tab:QuestionTemplates}
\end{table*}
}
\newcommand{\tablesocalresults}{
\begin{table*}[!t]
\centering
\resizebox{0.7\textwidth}{!}{
\begin{tabular}{lc|rrrrrrr}
\toprule
& \textit{SouthCal.} & Type 1 & Type 2 & Type 3 & Type 4 & Type 5 & Type 6 & Type 7  \\
\midrule
\multirow{3}{*}{\begin{turn}{90}BERT\end{turn}} & R@5 & 2.08 & 4.76  & 6.82 & 64.29  & 13.04 & 4.00 & 4.08   \\ 
& R@10 & 2.08 & 4.76  & 15.91 & 73.21 & 13.04 & 8.00 & 6.12  \\ 
& R@50 & 2.08 & 4.76   & 22.73 & 87.50 & 26.09 & 14.00 & 14.29  \\ 

\midrule
\multirow{3}{*}{\begin{turn}{90}GeoLM\end{turn}}  & 
R@5 & 4.17 & 19.05 &  4.55 & 67.86  & 26.09  & 10.00 & 10.20 \\ 
& R@10 & 4.17 & 26.19 &  6.82 & 75.00   & 30.43 & 16.00 & 12.24 \\ 
& R@50 & 4.17 & 38.10 &  20.45 & 82.14 & 43.48 & 22.00 & 20.41  \\ 

\bottomrule
\end{tabular}
}
\caption{Results on \textit{SouthCal} test dataset with DPR model using BERT and GeoLM as encoders respectively. The DPR model is trained on \textit{SouthCal} training set. }
\label{tab:result_socal}
\end{table*}
}

\newcommand{\tableilresults}{
\begin{table*}[!t]
\centering
\resizebox{0.7\textwidth}{!}{
\begin{tabular}{lc|rrrrrrr}
\toprule
& \textit{Illinois}  & Type 1 & Type 2 & Type 3 & Type 4 & Type 5 & Type 6 & Type 7  \\
\midrule
 \multirow{3}{*}{\begin{turn}{90}BERT\end{turn}}   & R@5 & 2.00 & 2.16 &  24.00 & 74.00  & 35.29 & 8.00 & 8.00 \\ 
& R@10 & 2.00 & 2.16 &  28.00 & 78.00  & 41.18 & 12.00 & 12.00 \\ 
& R@50 & 8.00 & 2.16 &  32.00 & 92.00  & 52.94 & 22.00 & 28.00 \\ 
 \midrule
 \multirow{3}{*}{\begin{turn}{90}GeoLM\end{turn}}  & 
R@5 & 0.00 & 2.16 &   26.00 & 62.00  & 64.71 & 4.00 & 8.00 \\ 
& R@10 & 2.00 & 2.60 &  28.00 & 68.00  & 64.71 & 10.00  & 16.00 \\ 
& R@50 & 10.00 & 4.76 &  32.00 & 88.00  & 64.71  &  22.00 & 32.00 \\ 
\bottomrule
\end{tabular}
}
\caption{Results on the unseen \textit{Illinois} dataset with DPR model using BERT and GeoLM as encoders respectively. The DPR model is trained on \textit{SouthCal} training set. Type 4 continues to stand out as a strong point for both models in zero-shot inference, with BERT achieving 92.00\% at R@50 and GeoLM achieving 88.00\%. This indicates that despite the region difference, both models generalize well to this type of question, likely because the semantic type knowledge transfers well from \textit{SouthCal} to \textit{Illinois}. }
\label{tab:result_il}
\end{table*}
}

\newcommand{\tablellmresult}{
\begin{table*}[!t]
\centering
\small
\resizebox{0.9\textwidth}{!}{

\begin{tabular}{l|rrrrrrrrr}
\toprule
\textit{Accuracy $\uparrow$} & Type 1 & Type 2 & Type 3 & Type 4 & Type 5 & Type 6 & Type 7 & Type 8 & Type 9 \\
\midrule

Gemini &  \textbf{42.68} & 0.00 & 6.10 & \textbf{100.00} & 12.73 & 10.37 & \underline{14.40} & 1.68 & 6.40  \\

GPT-3.5 &  35.98 & \textbf{6.98} & 4.27 & \textbf{100.00} & 5.45 & \underline{28.66} & \textbf{16.00} & \underline{2.23} & \textbf{96.80}\\

GPT-4o &  \underline{39.63} & 2.33 & \underline{7.93} & \textbf{100.00} & \textbf{29.09} & \textbf{37.20} & \underline{14.40} & \textbf{6.15} & \underline{91.60}  \\

LLaMA &  15.24 & \underline{4.65} & 4.88 & 88.41 & 14.55 & 22.56 & 5.60 & 1.12 & 8.40  \\

Mistral & 17.68 & 0.00 & \textbf{8.54} & \underline{99.39} & \underline{18.18} & 1.83 & 7.20 & 0.00 & 89.60 \\ 

\bottomrule
\end{tabular} }
\caption{The table presents the accuracy on the \textit{SouthCal} dataset using various LLMs in a Text-to-SQL task. \textbf{Bolded} scores indicate the highest performance for each question type. \underline{Underlined} scores denote the second-best results.}
\label{tab:result_llm_socal}

\end{table*}
}

\newcommand{\mapqans}{\textsc{MapQA}}
\newcommand{\mapqa}{\mapqans\xspace}

\title{\mapqans: Open-domain Geospatial Question Answering \\ on Map Data}

\author{Zekun Li}
\affiliation{%
  \institution{University of Minnesota, Twin Cities}
  \state{Minnesota}
  \country{USA}}
\email{li002666@umn.edu}

\author{Malcolm Grossman }
\affiliation{%
  \institution{University of Minnesota, Twin Cities}
  \state{Minnesota}
  \country{USA}}
\email{gros0562@umn.edu }

\author{Eric (Ehsan) Qasemi}
\authornote{This work was done while the author was a student at the University of Southern California (USC).}
\affiliation{%
 \institution{Oracle}
 \state{California}
 \country{USA}}
\email{ehsan.qasemi@oracle.com}

\author{Mihir Kulkarni}
\affiliation{%
  \institution{Pennsylvania State University}
  \state{Pennsylvania}
  \country{USA}}
\email{msk5743@psu.edu}

\author{Muhao Chen}
\affiliation{%
  \institution{University of California, Davis}
  \city{California}
  \country{USA}}
\email{muhchen@ucdavis.edu }

\author{Yao-Yi Chiang}
\affiliation{%
  \institution{University of Minnesota, Twin Cities}
  \city{Minnesota}
  \country{USA}}
\email{yaoyi@umn.edu}

\begin{document}

\renewcommand{\shortauthors}{Li et al.}

\begin{abstract} 

Geospatial question answering (QA) is a fundamental task in navigation and point of interest (POI) searches. While existing geospatial QA datasets exist, they are limited in both scale and diversity, often relying solely on textual descriptions of geo-entities without considering their geometries. A major challenge in scaling geospatial QA datasets for reasoning lies in the complexity of geospatial relationships, which require integrating spatial structures, topological dependencies, and multi-hop reasoning capabilities that most text-based QA datasets lack. To address these limitations, we introduce \mapqa, a novel dataset that not only provides question-answer pairs but also includes the geometries of geo-entities referenced in the questions. \mapqa is constructed using SQL query templates to extract question-answer pairs from OpenStreetMap (OSM) for two study regions: Southern California and Illinois. It consists of 3,154 QA pairs spanning nine question types that require geospatial reasoning, such as neighborhood inference and geo-entity type identification. Compared to existing datasets, \mapqa expands both the number and diversity of geospatial question types. We explore two approaches to tackle this challenge: (1) a retrieval-based language model that ranks candidate geo-entities by embedding similarity, and (2) a large language model (LLM) that generates SQL queries from natural language questions and geo-entity attributes, which are then executed against an OSM database. Our findings indicate that retrieval-based methods effectively capture concepts like \textit{closeness} and \textit{direction} but struggle with questions that require explicit computations (e.g., distance calculations). LLMs (e.g., GPT and Gemini) excel at generating SQL queries for one-hop reasoning but face challenges with multi-hop reasoning, highlighting a key bottleneck in advancing geospatial QA systems. The dataset is publicly available at \url{https://github.com/knowledge-computing/MapQA-dataset}

\end{abstract}
% refer to https://aclanthology.org/2022.naacl-main.320.pdf 
% and https://arxiv.org/pdf/2104.08712.pdf





\maketitle

\section{Introduction}

Open-domain question answering (QA) is a fundamental NLP task that aims to provide answers based on reference information from large corpora \citep{chen-etal-2017-reading}. Among the various types of questions an open-domain QA system addresses, geospatial questions stand out as a particular challenge. These questions require the system to ground and interpret geospatial entities (e.g., Minneapolis, the United Kingdom, Grand Canyon), reason about spatial quantities (e.g., distances, orientations), and understand spatial relationships (e.g., containment, adjacency, proximity, intersection) \citep{mai2021geographic}.

\begin{figure*}[t]
  \begin{center}
	\includegraphics[width=\linewidth]{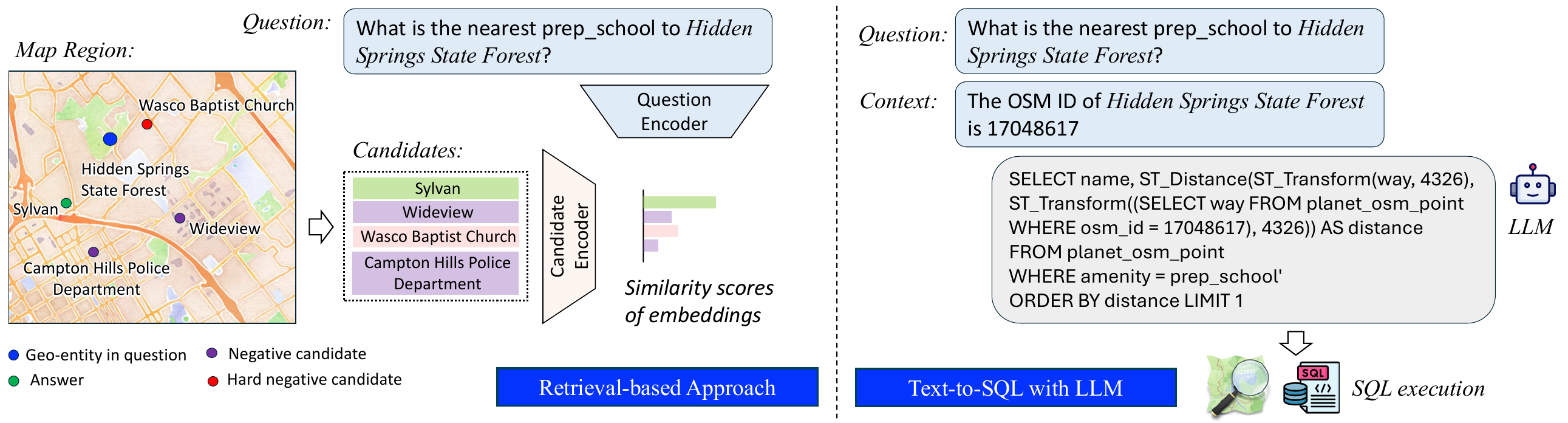}
  \end{center}
  \caption{
  The figure illustrates the two primary approaches we employed to tackle geospatial question answering (QA) problems. In the retrieval-based approach (e.g.,Dense Passage Retrieval~\cite{karpukhin-etal-2020-dense}), we construct the candidates by gathering all the entities in the geospatial database. QA is conducted by evaluating the similarity between the embeddings of the question and the candidate entities. In the large language model (LLM) text-to-SQL approach, the LLM generates SQL queries based on the given question and its context, where the context specifies attributes (e.g., OSM ID) of the query geo-entity. The QA process is conducted by running these SQL queries against the OpenStreetMap (OSM) database.
  }\label{fig:outline}
  % \vspace{-1em}
\end{figure*}

Despite their significance, geospatial questions remain underexplored in conventional open-domain QA benchmarks. Addressing them is crucial not only for enhancing natural language interfaces in Geographic Information Systems (GIS) and location-aware services but also for enabling advanced geo-analytical tasks, such as land cover classification and greenness density estimation \citep{scheider2021geo}. Moreover, robust geospatial QA systems enable interactive methods for users to access, acquire, and curate rich geospatial knowledge from maps, spatial databases, and spatially aware taskbots. Existing QA datasets primarily focus on general knowledge or domain-specific questions (e.g., Freebase database) \citep{rajpurkar2016squad, joshi-etal-2017-triviaqa, yang2018hotpotqa, kwiatkowski2019natural, berant-etal-2013-semantic, yang-etal-2015-wikiqa, choi-etal-2018-quac}, but they lack a dedicated focus on geospatial reasoning, such as spatial relationships. Tourism Questions ~\citep{contractor2019large} introduces a dataset for recommending geo-entities based on questions derived from tourism reviews. However, the dataset's entity type coverage is quite limited, containing only three categories: hotels, attractions, and restaurants. Contractor et al. \citep{contractor2021joint} construct a small-scale POI recommendation dataset designed to assess a model’s ability to infer distance relations. The dataset labels entity relations with three categories: (1) ``close to set X,'' (2) ``far from set X,'' and (3) a combination of both. However, these labels do not provide precise numerical distance constraints, making the spatial reasoning evaluation somewhat coarse-grained. There are also multi-modal datasets (e.g., MMMU \citep{yue2024mmmu} and MuirBench \citep{wang2024muirbench}) that include geographical questions as one of its disciplines. While the answers can be derived from input images such as satellite images, maps, or diagrams, the questions do not consistently focus on geo-entities, nor do they address geospatial relationships. The most relevant datasets to our work are GeoQuestions201 \citep{punjani2018template} and GeoQuestions1089 \citep{kefalidis2023benchmarking}, as they include questions involving semantic types, directional reasoning, and distance estimation. However, our dataset, \mapqa, differs in several key aspects. First, \mapqa is significantly larger, containing 3,154 QA pairs and covering 175 candidate geo-entity types, compared to GeoQuestions201 with 201 QA pairs and 28 geo-entity types (e.g., hotels, churches), and GeoQuestions1089 with 1,089 QA pairs. Second, \mapqa and GeoQuestions1089 include questions on both exact distance (e.g., within 50m) and relative distance (e.g., which is closer), whereas GeoQuestions201 contains only relative distance questions. Additionally, while both GeoQuestions201 and \mapqa are derived from the OpenStreetMap (OSM) database, GeoQuestions1089 is constructed from the YAGO2Geo knowledge graph. Finally, \mapqa leverages large language models (LLMs) to enhance the diversity and phrasing of questions (e.g., \textit{Which place is closer?} vs. \textit{Which location is nearer?}), simulating the various ways humans formulate queries in real-world scenarios. This linguistic variation increases the complexity of \mapqa, making it a more challenging benchmark for evaluating geospatial question-answering systems.

%We define geospatial questions as this subset of open domain QA that involve geographic entities, geographic features, or spatial relations. Recent work has shown remarkable promise for the construction of geographic question-answering datasets, but those datasets are either constructed using limited map information \citep{chang_mapqa_2022}, or lack the size to train traditional question-answering models \citep{punjani_template-based_2021}. To address the need for a large, generalized geographic question-answering dataset we propose \proska{dataset name here} as a new geographic question-answering dataset for open-domain QA.
%From the machine intelligence perspective, addressing geospatial questions extends the machine's capability for understanding large geospatial data, and conducting complex spatial and semantic reasoning on top of these.

% \begin{figure}[t]
%   \begin{center}
% 	\includegraphics[width=\linewidth]{figures/mapqa1.png}
%   \end{center}
%   \caption{Seven types of geospatial question included in \mapqa. Detailed statistics in \Cref{sec:statistics}. 
%   }\label{fig:mapqa}
%   % \vspace{-1em}
% \end{figure}

% \todo{ this gives a general overview but need to say more about why existing dataset does not provide solution (entirely or partially to this problem). the abstract talks about the existing work limits in qa quantity and geocoverage. are improving the qa quantity and geocoverage help address the problem said here? or is the varieety (e.g., 9 types) important? }

To advance research on geospatial QA, we introduce \mapqans, a novel open-domain QA benchmark as our \textit{first} contribution. Unlike traditional open-domain QA tasks, where systems retrieve and interpret semantically relevant textual passages, \mapqa relies on extensive geospatial map data as reference information. To answer questions in \mapqa, a QA system must identify reference map data that is both spatially and semantically relevant to the query. Moreover, these questions often require reasoning chains that incorporate both semantic and geospatial relationships. For instance, the question ``What is the closest bus stop to Starbucks Coffee?'' is highly context-dependent, given the presence of numerous distinct ``Starbucks Coffee'' locations~\citep{mai2019unanswerable}. A robust geospatial QA dataset should encompass diverse question types, including proximity, directionality, and the semantic classification of geo-entities. Additionally, the dataset must incorporate questions that require entity disambiguation based on context (e.g., differentiating between multiple ``Starbucks'' locations), as well as questions with multiple plausible answers (e.g., the amenity type of ``Northwestern Hospital'' could be classified as both a \textit{facility} and \textit{healthcare}). \Cref{tab:QuestionTemplates} outlines the nine distinct question types included in the dataset. Each question type is designed to encompass one or multiple steps of reasoning, covering various spatial concepts, such as amenity types and distances, which allows for a comprehensive evaluation of geospatial reasoning.

\tabledataset

Beyond introducing the \mapqa dataset, our second contribution is a systematic benchmarking of two families of methods tailored to the \mapqa problem setting. The first family focuses on neural retrieval. In addition to evaluating standard dense retrievers such as DPR \citep{karpukhin2020dense}, we explore variants that integrate a geospatially grounded language model \citep{li-etal-2023-geolm}, which jointly encodes textual descriptions and geo-coordinates to enhance geospatial reasoning. This retrieval-based approach ranks candidate geo-entities by comparing their embedding similarities to the given questions. The second family of methods is based on text-to-SQL generation, where large language models (e.g., GPT, LLaMA) translate natural language questions into executable PostgreSQL queries \citep{dong2023c3}. These queries are then executed using the PostGIS engine to retrieve answers from a geospatial database. However, this approach relies on prior knowledge of the database schema, including table and column names; otherwise, incorrect or randomly generated attribute names can significantly impact query executability.

% \muhao{question: do we combine these two in the same way as DeCaF does? (first seq2sql, then DPR if seq2sql fails to get anything)}

In \mapqa, we use Southern California and Illinois as sample study regions to highlight the challenges of geospatial reasoning. The Southern California region includes 606,773 entities across 621 amenity types, while the Illinois region encompasses 92,415 entities across 175 amenity types. \mapqa consists of 3,154 question-answer pairs, capturing nine types of geographical relations, with the aforementioned ~699K geo-entities and ~700 amenity types serving as candidate answers. The process for generating question-answer pairs can be easily extended to other regions, and we provide a script for generating additional data. Experimental results show that retrieval-based models can associate amenity types with geo-entities and understand the closeness and direction concepts although it has limitations on predicting the exact values of distances. (This limitation arises from the fundamental difference between distance, which is a continuous variable, and retrieval-based approaches, which are inherently designed to select from a discrete set of candidate entities. As a result, distance calculation cannot be directly formulated as a retrieval-based problem, making precise numerical estimation challenging for such models.) Additionally, with moderate fine-tuning, this approach can perform reasonably well on unseen study regions. The second approach involves using large language models (LLMs) to generate SQL queries based on the questions and geo-entity attributes, which are then executed against the OSM database. Our findings suggest that LLMs, such as GPT and Gemini, are very strong at generating SQL queries that only require one-hop reasoning. 

%The main contributions we propose are: (1) We address the need for a large, open-domain geographic question-answering dataset for training and evaluating geographic question-answering models; (2) We develop two new geographic question-answering models that leverage spatial information to answer geographic questions and show \proska{re-iterate significant results shown by models.}

\section{MapQA Dataset}
\subsection{Problem Statement}

% Recent advancements have focused on representing geometries and spatial relationships using machine learning models and analyzing them in conjunction with text. However, there is a notable lack of large-scale datasets that combine question-answering with the representation of spatial objects. 
Our motivation for creating this dataset is to address the gap in question-answering datasets that necessitate spatial reasoning to accurately answer questions. Geospatial question answering typically requires multi-hop reasoning, and \mapqa is designed to simulate real-world scenarios that demand an understanding of distance, direction, and geo-entity types. For instance, to answer the question, \textit{What are the bars within 50m of South Park Brewing?}, the model must first retrieve the geo-entities close to South Park Brewing, then infer the type of these entities and exclude non-bar geo-entities. Additionally, we include questions that require only single-hop reasoning, such as ``What is the amenity type of Chase?" to assess whether models can effectively learn and apply semantic type information.

 The \mapqa dataset includes three types of answers: 1) geo-entity names (e.g., \textit{What restaurant is adjacent to Luther Burbank Savings?}) and 2) amenity types (e.g., \textit{What amenity is available at Port Police?}) 3) distances (e.g., \textit{What is the distance between Brentwood and Coquitlam?}). For the first two types, we collect all the OSM geo-entities and amenities in the study regions as candidate answers for the retrieval-based model, and the evaluation is conducted by comparing whether the correct geo-entity or amenity type is retrieved. For the third type, we execute the human-verified SQL queries to obtain the ground-truth distance values. The evaluation is done by calculating the distance difference between the ground truth and prediction. If the offset is smaller than 100 meters, we treat the prediction as a correct answer. 

\subsection{Data Collection}
To derive question-answer pairs, we harness the wealth of structured geospatial data available on OpenStreetMap (OSM).
OpenStreetMap (OSM) is a freely accessible geographic database that is continually updated and managed by a dedicated community of volunteers through open collaboration. 
OSM is renowned for its \textit{crowd-sourced} geospatial information with \textit{wide} geographical coverage, making it an invaluable primary data source for our task. OSM provides meticulously curated details about diverse amenities, geographic landmarks, and geographic coordinates.

To generate question-answer pairs from the OSM data, we adopt a framework that deviates from traditional human-supervised methods by enabling the acquisition of labeled data in a more expedient and cost-effective manner while maintaining acceptable margins of error. 
Instead of relying on an army of trained annotators, the domain experts craft heuristic labeling functions (LFs) to repurpose the validated structured knowledge into a format suitable for the secondary task \cite{ratner2016data}, which is question-answering in our case.
Consequently, instead of engaging a substantial workforce of human annotators for question generation, the primary challenge lies in designing these heuristics to query the existing knowledge and transform the results into a suitable format for the task at hand.

We employ a curated set of nine question templates, summarized in \Cref{tab:QuestionTemplates}, each designed to generate diverse question-answer instances and assess various aspects of spatial reasoning. The spatial concepts underlying these questions are listed in the last column. To enhance linguistic diversity, we leverage LLMs (e.g., GPT), to generate paraphrased variations of each template by modifying phrasing (e.g., replacing \textit{is adjacent to} with \textit{neighbors}). This approach increases dataset variability, improving its robustness for training and evaluating models across a broader range of natural language scenarios.

\begin{itemize}

  \item \textbf{Attribute Association} is essential for tasks such as service recommendations (e.g., locating nearby gas stations or restaurants) and in understanding the function or purpose of geographic features. Questions like ``What amenity is available at [location]?'' and ``What are the [amenities] within 50m of [location]?'' evaluate the model's ability to link specific attributes or amenities to particular locations. The former asks for the amenity type as the answer, while the latter requires identifying the geo-entity itself.

  \item \textbf{Spatial \& Directional Proximity}: Understanding the proximity of geo-entities is fundamental for applications such as routing, navigation, and urban planning. Our work emphasizes the importance of both spatial and directional proximity between geo-entities. For example, we assess whether models can determine which of two candidates is closer to a given location (e.g., ``Which is closer to ChargePoint, [location A] or [location B]?'') and whether the model can perform directional inference (e.g., What is the nearest [amentity] \underline{\textit{west}} of [location]?) Additionally, we extend this evaluation to include line geometries such as streets, constructing questions like ``Which [amenity] is nearest to the \underline{\textit{junction}} of [location A] and [location B]?'' 

  \item \textbf{Distance Calculation}: Accurate distance measurement between geo-entities is essential for tasks such as recommending optimal routes, evaluating accessibility, and defining service areas. Questions like ``How far is [location A] from [location B]?'' directly require distance calculations between two geographic points, with the answer being the precise distance value. 

\end{itemize}

To limit the size and scope of the data extracted from OSM, we focused on the Southern California and Illinois regions. \footnote{Data dump downloaded on Sep 18th, 2023, from \url{https://download.geofabrik.de/north-america/us/california/socal-latest.osm.pbf}} For instance, the process for generating instances using the template \textit{"What \{amenity\} is adjacent to \{location\_name\}?"} begins by establishing a connection to the database and selecting a random amenity type and location name from a predefined pool as seeds. This ensures that the dataset contains a diverse range of location-based queries. Next, for each selected location, the SQL code queries the database to find all amenities of the given type that are within 50 meters of the reference location using \texttt{PostGIS} spatial functions. To avoid redundant entries, the reference location itself is excluded from the results. If no matching amenities are found within the specified radius, the location is skipped to maintain meaningful question-answer pairs after reaching 10 empty amenities. Finally, the extracted data is formatted into structured outputs such as \texttt{CSV} and \texttt{JSON}, facilitating its integration into geospatial question-answering datasets. This approach leverages OSM’s structured geospatial knowledge and \texttt{PostGIS} spatial indexing to efficiently generate diverse, high-quality questions while maintaining spatial relevance.

\subsection{Dataset Statistics} \label{sec:statistics}

\begin{figure*}[t]
  \begin{center}
	\includegraphics[width=\linewidth]{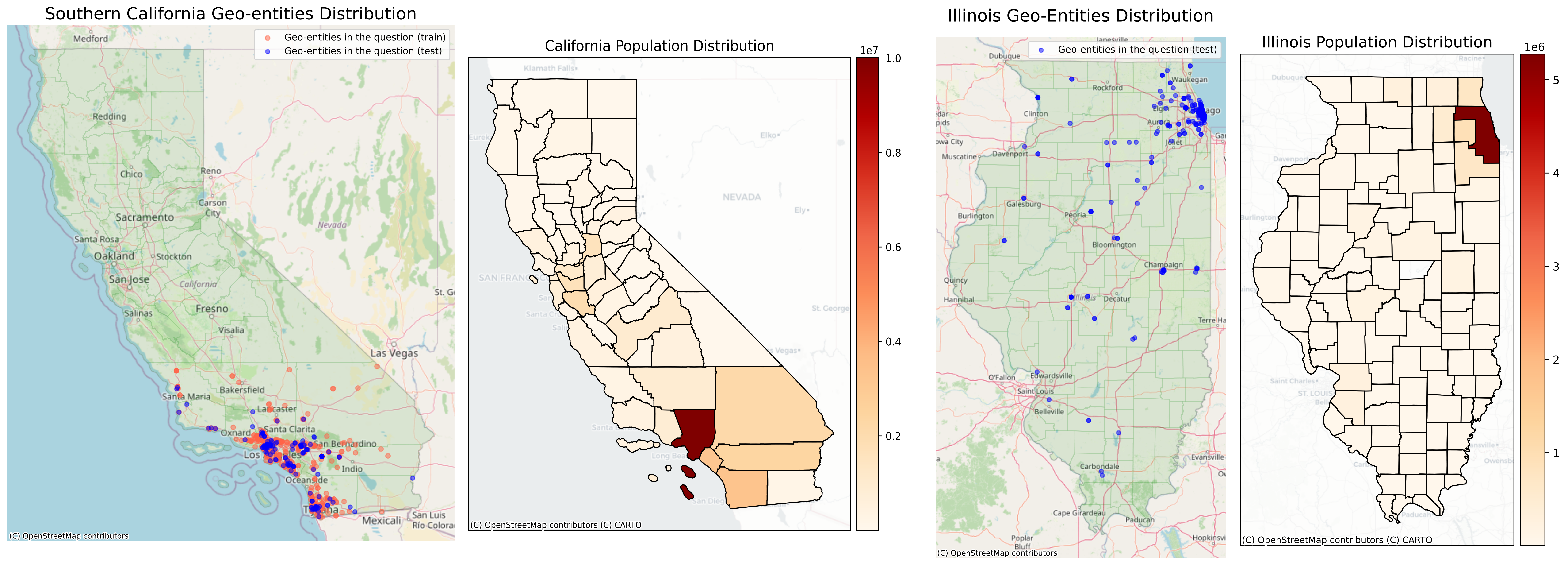}
  \end{center}
  \caption{
    The figure illustrates the distribution of geo-entities in the questions for two study regions. Orange dots represent geo-entities from the training set, while blue dots denote those in the test set. Since Illinois serves as a zero-shot test set, it contains no training geo-entities (i.e., no orange dots). The \textit{SouthCal} split includes only geo-entities from Southern California, rather than the entire state, to optimize computational efficiency. However, the same methodology is directly applicable to Northern California or other regions. The geo-entity distribution closely follows county-level population patterns derived from US Census data, indicating that the sampling strategy effectively captures real-world user interests and reflects typical spatial distributions in practical applications. }\label{fig:entity-distribution}

\end{figure*}

\begin{figure}[t]
  \begin{center}
	\includegraphics[width=\linewidth]{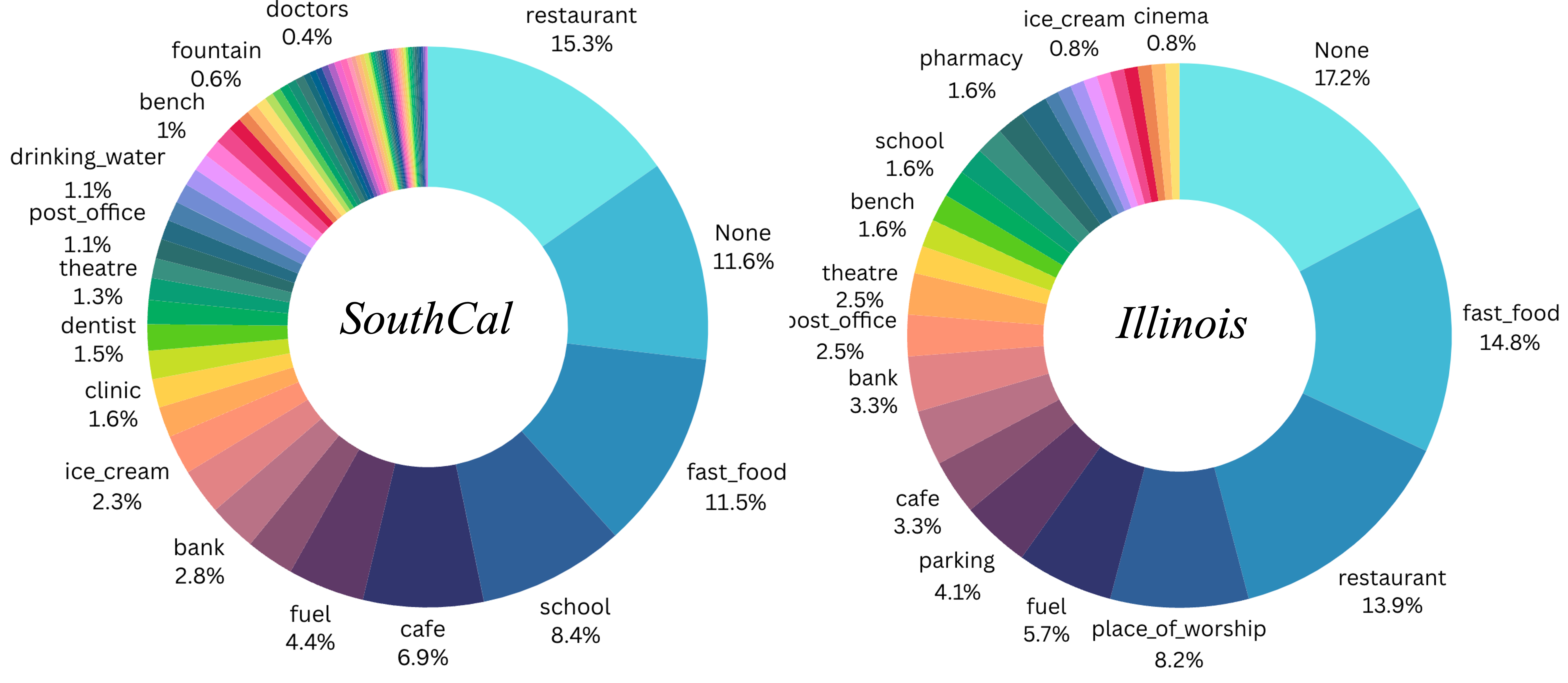}
  \end{center}
  \caption{
    The figure shows the amenity type distribution in the answers for \textit{SouthCal} and \textit{Illinois} study region, with \textit{SouthCal} featuring 73 distinct amenity types and \textit{Illinois} consisting of 29. During the training of retrieval-based models, a total of 175 amenity types were used as candidate options. Due to space constraints, the pie chart does not include all amenity types in its labels. For a comprehensive list of the amenity types referenced in the answers, please refer to \Cref{ap:amenity type}.
  }\label{fig:amenity-distribution}

\end{figure}

The final dataset comprises 3,154 question-answer pairs, categorized by type as detailed in \Cref{tab:QuestionTemplates}. For the Southern California (\textit{SouthCal}) region, which includes 2,206 pairs, 80\% are allocated for training and 20\% for testing. During model training, the training set is further divided into \textsc{train} and \textsc{val} subsets using an 80/20 split. This dataset includes 644 geo-entities in the training questions and 162 in the test questions.  The \textit{Illinois} region serves as the zero-shot evaluation set, containing 948 QA pairs and 348 geo-entities. \Cref{fig:entity-distribution} visualizes the geographical distribution of geo-entities across both study regions, and \Cref{fig:amenity-distribution} presents the amenity type distribution. 

% The test set (296) comprises 197 with geo-entities as answers and 99 with amenities as answers. 

% \subsection{Geographic Information}

\section{Baseline Methods for MapQA} 
\label{sec:method}
We developed four baseline models to evaluate the dataset, three use retrieval-based QA methods derived from DPR and the last model uses Large Language Models (LLM) in a text-to-SQL setup.

\subsection{Dense Passage Retrieval}
Dense Passage Retrieval (DPR) is a retrieval-based model designed for open-domain question answering. It works by encoding text passages using a passage encoder and storing them in a corpus. When a query is received, a question encoder generates its embedding, which is then used to retrieve the most relevant passages from the corpus. In our adaptation of DPR, we repurpose the passage encoder to encode geo-entity names (or amenity types) as candidate answers, rather than textual passages. For the \textit{SouthCal} region, we construct a candidate pool comprising over 600,000 geo-entities and more than 600 amenity types. During training for geo-entity recommendation, positive contexts correspond to the correct geo-entity answers, while negative contexts are derived from random samples within the candidate pool, supplemented by \textit{hard negatives}, which consist of neighboring geo-entities in close proximity to the target entity. During retrieval, the candidate pool serves as the corpus, and its feature embeddings are used to retrieve the $k$ nearest candidates relevant to the query. For amenity recommendation questions, the procedure remains similar, except that negative contexts consist solely of randomly sampled amenities from the pool. In the \textit{Illinois} region, the geo-entity pool includes over 92,000 candidates, while the amenity pool contains 175 categories.

In addition to the BERT encoder, we also explore using a geospatial language model, namely GeoLM\cite{li-etal-2023-geolm} as the encoder. The primary difference between base DPR and the DPR-GeoLM is that GeoLM leverages spatial information to aid in encoding questions and passages. GeoLM encodes the spatial information derived from the latitude and longitude of spatial entities in a sentence. Specifically, GeoLM uses the tokenized text and a constructed coordinate sentence of the same length as input. The embeddings of the text sentence and the latitude and longitude sentences are separately calculated using then combined to create the final embedding. It should be noted that GeoLM only supports representing point geometries, making their use in geographic question answering limited to those types of geometries. Work has been done towards embedding systems that can represent all geometries \citep{mai2023towards} but none have applied this towards natural language QA.

\subsection{Text-to-SQL}
In the Text-to-SQL models, we use multiple Large Language Model, e.g. GPT-4 \cite{openai2023gpt}, LLaMA-3\cite{touvron2023llama} and Gemini \cite{team2023gemini}, to generate SQL queries to extract the answer given the natural language prompt containing the original question plus general information ( \Cref{tab:ttsql_prompt}). We then run the generated SQL on the server, extract the results, and compare them with the ground truth. 

\tableprompt

To generate the natural language prompt, we employed the template outlined in \Cref{tab:ttsql_prompt}. This template defines the structure of the OSM table and provides essential context, such as the OSM-ID associated with a given geo-entity. The template consists of three main sections:

\begin{enumerate}
    \item \textbf{Table Schema and Structure:} The first section offers a comprehensive description of the "planet\_osm\_point" table, including its columns, column types, and a brief explanation of the data stored in each column. It also provides general contextual information about the table and additional insights into each column’s role.
    
    \item \textbf{Natural Language Question:} The third section presents the question in natural language, without any modifications.
    
    \item \textbf{Contextual Information:} The final section generates additional context for the model, formatted as a series of sentences following the template: “The OSMID of <location\_name> is <osm\_id>.” This contextual information serves to assist the model in effectively formulating queries by providing access to actual OSM identifiers.
\end{enumerate}

We selected large language models (LLMs) over dedicated Text-to-SQL models for two primary reasons. First, while existing Text-to-SQL models, such as LSTM-based approaches (e.g., Seq2SQL \cite{zhong2017seq2sql}, SQLNet \cite{xu2017sqlnet}) and transformer-based models (e.g., BERT-to-SQL \cite{wang2019rat, lin-etal-2020-bridging, deng-etal-2021-structure}, T5-to-SQL \cite{xie-etal-2022-unifiedskg}), excel at parsing standard SQL queries (e.g., \texttt{SELECT}, \texttt{WHERE}) and more advanced aggregation clauses (e.g., \texttt{GROUP BY}), they struggle with geospatial queries that require parsing geometries and generating PostGIS functions such as \texttt{ST\_WITHIN}, \texttt{ST\_X}, \texttt{ST\_Y}, and \texttt{ST\_Transform}. As a result, these models would require substantial fine-tuning to work effectively with the \mapqa dataset. Second, LLMs are trained on extensive, diverse text data, which enhances their ability to understand complex queries and generate accurate geospatial statements. Their flexibility allows them to adapt to new tasks and input prompts, making them a better choice for the \mapqa task.

\tablesocalresults
\tableilresults
% \tabletrainratio

\section{Experiments}

In this section, we benchmark the performance of the two groups of models outlined in \Cref{sec:method}. The first group consists of well-established retrieval backbones, specifically Dense Passage Retrieval (DPR) models, using two encoder architectures: BERT and GeoLM. The second group includes a range of large language models (LLMs) applied in a Text-to-SQL setup. For the LLMs, we benchmarked GPT-3.5-turbo, GPT-4o \citep{openai2023gpt}, Gemini-1.5-flash \citep{team2023gemini}, LLaMA-3-8B-Instruct \citep{dubey2024llama}, and Mistral-7B-Instruct-v0.3 \citep{jiang2023mistral}.

\subsection{Dense Passage Retrieval}
\textbf{Training: } We train two Dense Passage Retrieval (DPR) variants, DPR-BERT and DPR-GeoLM, on the \textit{SouthCal} dataset, and evaluate their performance in two distinct settings: (1) an \textbf{in-region} evaluation, where the models are tested on a hold-out set from the Southern California region, and (2) a \textbf{cross-region} evaluation, where the models are tested in a zero-shot manner on the Illinois region. The questions are further categorized into two types based on their expected output: (1) questions where the output is a geo-entity and (2) questions where the output is an amenity name. To address these differences, we train separate models for each output type using both DPR-BERT and DPR-GeoLM. However, DPR-based models face two key limitations: (1) difficulty in resolving questions involving geo-entities with line geometries, and (2) challenges in handling questions where the output is a distance value, as these cannot be framed as simple retrieval problems. Due to these constraints, Type 8 and Type 9 questions are excluded from the DPR experiments.

% \tabledprstats

\tablellmresult

\noindent \textbf{Candidate Construction: } Following the approach outlined in \citep{karpukhin2020dense}, we employ positive and negative sampling techniques for candidate generation during training. For each question, we create positive, negative, and hard-negative candidates. Positive candidates are selected based on BM25 scores, identifying the highest-ranked geo-entity that corresponds to the question name. In contrast, negative and hard-negative candidates are generated using a combination of three methods: (1) Random: a randomly chosen candidate from the geospatial database; (2) BM25: the top candidates retrieved by BM25 that do not provide the correct answer but align with the most question tokens; and (3) Gold: geo-entity answers from other questions present in the training set. Hard-negative candidates are specifically identified using method (2), where we select candidates with the highest BM25 scores relevant to the input question. To compile the negative candidates, we utilize both methods (1) and (3): method (1) contributes random candidates that are not already included in the positive or hard-negative lists, while method (3) provides candidates that are positive answers from different questions within the training set, which are repurposed as negatives. In line with the strategy of \citep{karpukhin2020dense}, we implement in-batch negative training, allowing us to select gold negative candidates from other questions contained within the same batch. This technique offers a straightforward and memory-efficient means of reusing candidates already present in the batch. Additionally, method (1) ensures the extraction of random candidates that do not overlap with those in the positive and hard-negative candidate lists, further enriching the negative candidate pool.

% Dense Passage Retrieval (DPR) tackles the task of open-domain QA by mapping any text passage to a d-dimensional vector using a dense encoder. It builds an index of M passages that are used for retrieval. On run-time, DPR applies the encoder, and maps input questions to a d-dimensional vector and retrieves the top k passages closest to the encoded question. Similarity is defined by the dot product of their vectors. 

% In this experiment setup, we use dual BERT (base, uncased) encoders for the passage and query encoder. Positive samples are selected from the collection of labels for each query and negative samples are selected from passages with high BM25 scores that are not in the positive set. \\
% Each passage and question is tokenized and padded to the vector size of d = 768.\\
% During training questions and their related positive and negative passages are encoded using the dual encoder setup. The dot product is calculated between each passage and the query and negative log-likelihood loss is calculated. \\

% \tablellmerror

% \tablelaccuracyllm
% \tablelerrorllm

\begin{figure}[t]
  \begin{center}
	\includegraphics[width=\linewidth]{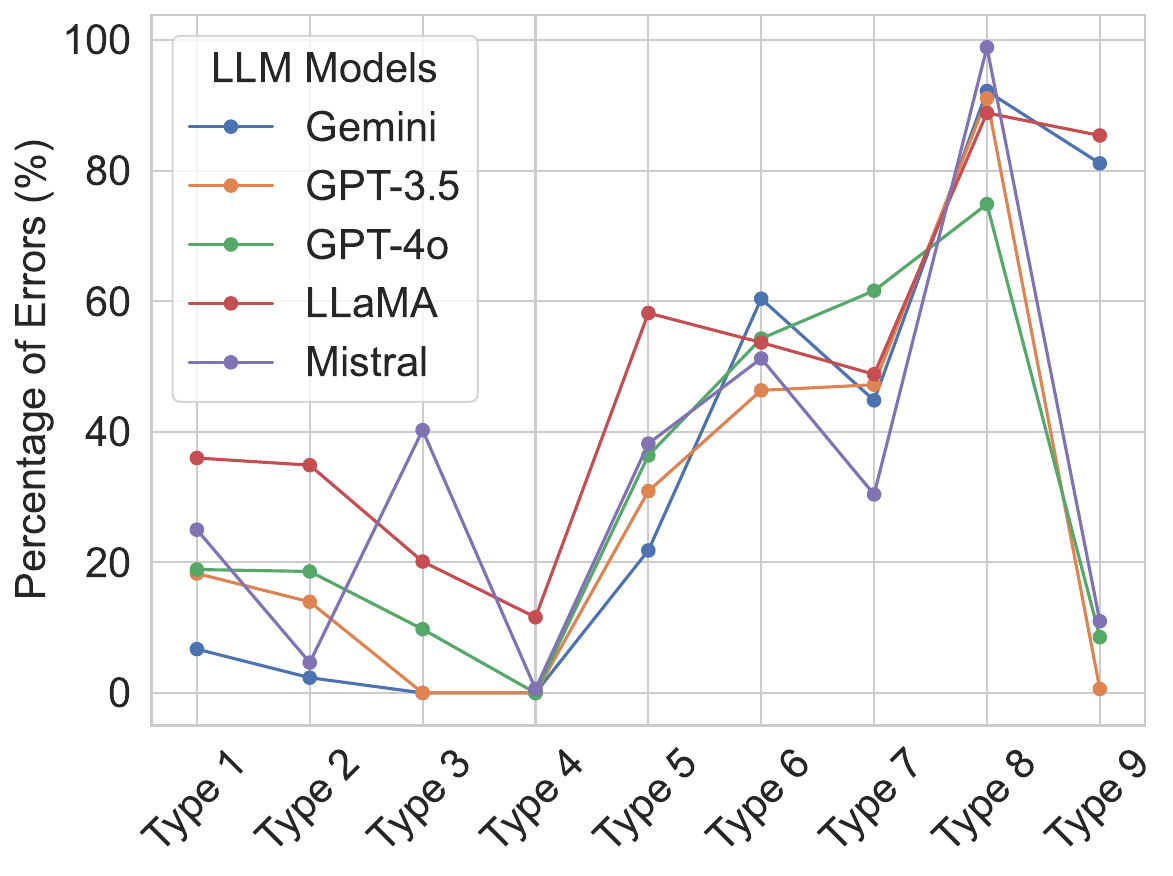}
  \end{center}
  \caption{
  The percentage of malformed SQL scripts (e.g., those containing syntax errors) generated by each LLM in the Text-to-SQL experiments.
  }\label{fig:llm_error}
  \vspace{-2em}
\end{figure}

\noindent \textbf{Results:} \Cref{tab:result_socal} shows the results in the first setting -- in-region evaluation. DPR-BERT and DPR-GeoLM demonstrate varying performance across different question types. For Type 4 questions, both encoders perform exceptionally well, with DPR-BERT achieving 87.50\% (R@50), and DPR-GeoLM reaching 82.14\%. This suggests that both encoders have the ability to associate the semantic type (e.g., amenity type) information with the geo-entities. GeoLM outperforms BERT across most other question types, especially in Type 2 (R@50: 38.10\% vs. 4.76\%), Type 3 (R@50: 20.45\% vs. 22.73\%), and Type 5 (R@50: 43.48\% vs. 26.09\%). This highlights GeoLM’s superior capability in understanding exact distance concepts (e.g., 50m, 100m). Performance on Type 1 questions remains low for both DPR-BERT and DPR-GeoLM, with DPR-BERT consistently scoring 2.08\% across all recall metrics, while DPR-GeoLM shows marginal improvement at R@5 and R@10 (4.17\%), though the scores remain relatively low. This suggests that both models face challenges with Type 1 questions, which require correctly inferring adjacent concepts and amenity types simultaneously. For the cross-region evaluation on the zero-shot \textit{Illinois} dataset (see \Cref{tab:result_il}), the results indicate relatively stable performance compared to the in-domain \textit{SouthCal} results. This stability suggests that retrieval models, once trained to encode spatial information within a particular region, can generalize to unseen geographic regions. Overall, DPR-GeoLM demonstrates more consistent performance across a broader range of question types, highlighting its robustness in geospatial question answering.

\subsection{Text-to-SQL Using LLMs}

To benchmark the performance of Text-to-SQL models on the \mapqa dataset, we report accuracy as the evaluation metric, which represents the percentage of cases where the generated answer matches the ground-truth data. To account for and distinguish cases of malformed SQL (e.g., syntax errors), we also report the percentage of such instances.

\noindent \textbf{Results:} \Cref{tab:result_llm_socal} presents the accuracy results of various Text-to-SQL models on the \textit{SouthCal} dataset. Type 4 stands out as a category where all models, except LLAMA, perform at or near the maximum possible score, with Gemini, GPT-3.5, and GPT-4o all achieving 100.00\% and Mistral close behind at 99.39\%. This high performance can be attributed to the relatively simpler one-hop reasoning required for generating SQL queries in Type 4, making it easier for LLMs to handle. Both GPT-3.5 and GPT-4o consistently outperform other models across multiple question types, with GPT-4o excelling particularly in Type 5 (29.09\%) and Type 6 (37.20\%). These question types involve more complex spatial predicates, such as \texttt{ST\_X}, \texttt{ST\_Y}, and \texttt{ORDER\_BY}, highlighting GPT-4o's superior ability to manage intricate SQL query generation. However, Type 8 and Type 9 emerge as the most challenging categories, with most of the models struggling to produce accurate results. Our analysis of malformed SQL queries, shown in \Cref{fig:llm_error}, indicates that these types have the highest error rates across all categories. This suggests that generating the correct syntax for finding intersections between two geo-entities and calculating distances remains a significant challenge for current LLMs. These errors emphasize the ongoing difficulties in handling complex spatial relationships and advanced geospatial reasoning in Text-to-SQL tasks.

% \subsection{Error Analysis}

% Candidates:
% \begin{itemize}
%     \item detailed performance. e.g. per question type, topology, multi-hop vs single-hop, no-answer question
%     \item transfer learning, e.g. question types, different locations
%     \item error analysis. Similar to the one Mihir did
% \end{itemize}

\section{Related Work}
% \proska{possibly also text-to-sql and retrieval models}
% \proska{Seq2SQL, RAT-SQL, SQLNet}

% talk about QA datasets

\subsection{Geospatial NLP}

In recent years, researchers have increasingly applied natural language processing (NLP) models in the geospatial domain. Numerous named entity recognition (NER) models \citep{dernoncourt2017neuroner, devlin-etal-2019-bert, zhuang-etal-2021-robustly} have been adapted to tackle toponym detection \citep{halterman2017mordecai, tao2022geographic} and toponym linking \citep{grover2010use, weissenbacher2019semeval, gritta-etal-2017-vancouver, gritta-etal-2018-melbourne}. General-purpose NLP models have also been applied to geospatial relation extraction \citep{yu2015bootstrapping}. For example, \citet{yang2022spatial} designed a network based on the BERT architecture to classify geo-entity relations into 14 distinct categories. Additionally, points of interest (POI) recommendation and prediction tasks, such as store recommendation and house price prediction \citep{gao2022geobert}, have been addressed using fine-tuned language models. However, these models typically consider only the linguistic context of toponyms (i.e., place names), overlooking critical geolocation data and spatial correlations with neighboring entities. More recently, \citet{li2022spabert} trained a language model that captures geospatial context for geo-entity feature representation and later extended this work to geospatially grounded natural language understanding \citep{li-etal-2023-geolm}. In this paper, we demonstrate that NLP models can effectively perform geospatial QA tasks, achieving performance on par with domain-specific models.

\subsection{Open Domain QA}
Open-domain question answering (QA) has been a prominent area of research in natural language processing (NLP) and information retrieval (IR), with the goal of developing systems that can retrieve relevant information from knowledge sources to answer user queries without being constrained by predefined domains. Over the past decade, numerous QA benchmarks have been released, reflecting the growing interest in advancing open-domain QA systems. Notable datasets, such as HotPotQA \citep{yang2018hotpotqa}, TriviaQA \citep{joshi-etal-2017-triviaqa}, WebQA \citep{chang2022webqa}, and SearchQA \citep{dunn2017searchqa}, serve as crucial resources for evaluating QA models. These datasets span a wide range of topics and question types, allowing researchers to assess model generalization across various domains. However, a significant limitation of these benchmarks is their lack of focus on geospatial questions, which present unique challenges in open-domain QA. Geospatial questions, which involve queries about locations, distances, and spatial relationships, present distinct challenges for open-domain question-answering systems. This paper addresses these challenges by developing an open-domain question-answering system specifically designed for geospatial queries.
\section{Conclusion and Future work}
In this work, we address the unique challenges of geospatial question answering (QA) by introducing \mapqa, a novel dataset derived from OpenStreetMap (OSM), consisting of 3,154 question-answer pairs across two diverse regions: Southern California and Illinois. Our dataset features a wide variety of geospatial reasoning tasks, making it a robust benchmark for evaluating QA systems in this domain. We also explored two primary approaches, 1) retrieval-based methods and 2) LLM-based text-to-SQL generation, to tackle the geospatial QA challenges presented by \mapqa. Overall, \mapqa represents a significant step forward in geospatial QA research, providing a challenging and comprehensive benchmark that can spur further advancements in the field. In the future, we hope to expand geospatial relations to include topological relationships, such as containment, touching, and overlapping.

\newpage

\bibliography{anthology, anthology_p2, custom}

\bibliographystyle{ACM-Reference-Format}

\appendix
% \newpage

\section{Appendix} \label{ap:amenity type}

\Cref{tab:amenity-distribution-cal} and \Cref{tab:amenity-distribution-il} show the full list and occurrence count of amenity types in the answers.

\begin{table}[ht]
\centering
\begin{tabular}{|l|c|| l|c|}
\hline
\textbf{Category} & \textbf{Count} & \textbf{Category} & \textbf{Count}  \\
\hline
restaurant & 121 & kindergarten & 3 \\
None & 92 & doctors & 3 \\
fast\_food & 91 & telephone & 3 \\
school & 67 & townhall & 3 \\
cafe & 55 & bicycle\_rental & 3 \\
fuel & 35 & studio & 2 \\
place\_of\_worship & 22 & social\_centre & 2 \\
bank & 22 & marketplace & 2 \\
fire\_station & 21 & ferry\_terminal & 2 \\
ice\_cream & 18 & car\_rental & 2 \\
bar & 14 & family\_planning & 1 \\
clinic & 13 & psychic & 1 \\
parking\_entrance & 13 & university & 1 \\
dentist & 12 & college & 1 \\
bicycle\_parking & 11 & driving\_school & 1 \\
theatre & 10 & parcel\_locker & 1 \\
pharmacy & 9 & cinema & 1 \\
post\_office & 9 & bbq & 1 \\
library & 9 & bear\_box & 1 \\
drinking\_water & 9 & bicycle\_repair\_station & 1 \\
atm & 9 & public\_transport\_tickets & 1 \\
vending\_machine & 8 & reception\_desk & 1 \\
pub & 8 & bureau\_de\_change & 1 \\
bench & 8 & nursing\_home & 1 \\
toilets & 8 & public\_bath & 1 \\
post\_box & 6 & charging\_station & 1 \\
parking & 5 & watering\_place & 1 \\
fountain & 5 & events\_venue & 1 \\
dojo & 5 & music\_school & 1 \\
community\_centre & 4 & waste\_basket & 1 \\
clock & 4 & food\_court & 1 \\
social\_facility & 4 & arts\_centre & 1 \\
veterinary & 4 & bus\_station & 1 \\
prep\_school & 4 & courthouse & 1 \\
police & 3 & prison & 1 \\
car\_wash & 3 & grave\_yard & 1 \\
trailer\_park & 1 & & \\

\hline
\end{tabular}
\caption{Amenity type count in the answers for \textit{SouthCal} study region.}
\label{tab:amenity-distribution-cal}
\end{table}

\begin{table}[ht]
\centering
\begin{tabular}{|l|c|}
\hline
\textbf{Category} & \textbf{Count} \\
\hline
None & 21 \\
fast\_food & 18 \\
restaurant & 17 \\
place\_of\_worship & 10 \\
fuel & 7 \\
parking & 5 \\
cafe & 4 \\
grave\_yard & 4 \\
bank & 4 \\
post\_officer & 3 \\
theatre & 3 \\
post\_box & 2 \\
bench & 2 \\
fountain & 2 \\
school & 2 \\
dentist & 2 \\
clinic & 2 \\
pharmacy & 2 \\
ferry\_terminal & 2 \\
atm & 1 \\
nightclub & 1 \\
bicycle\_rental & 1 \\
ice\_cream & 1 \\
bar & 1 \\
social\_centre & 1 \\
bicycle\_parking & 1 \\
childcare & 1 \\
parking\_entrance & 1 \\
cinema & 1 \\
\hline
\end{tabular}
\caption{Amenity type count in the answers for \textit{Illinois} study region.}
\label{tab:amenity-distribution-il}
\end{table}


\end{document}